\documentclass[letterpaper]{article} 
\usepackage{aaai2026}  
\usepackage{times}  
\usepackage{helvet}  
\usepackage{courier}  
\usepackage[hyphens]{url}  
\usepackage{graphicx} 
\usepackage{natbib}  
\usepackage{caption} 
\usepackage{algorithm}
\usepackage{algorithmic}
\usepackage{enumitem}
\setlist[enumerate]{label=\arabic*.}

\usepackage{amsmath}
\usepackage{array}
\usepackage{booktabs}
\usepackage[most]{tcolorbox}

\usepackage{newfloat}
\usepackage{listings}
\DeclareCaptionStyle{ruled}{labelfont=normalfont,labelsep=colon,strut=off} 
\floatstyle{ruled}
\newfloat{listing}{tb}{lst}{}
\floatname{listing}{Listing}
\usepackage{csquotes}  
\usepackage{epigraph}
\newcommand{\alias}{ARCHE Bench}

\title{ARCHE: A Novel Task to Evaluate LLMs on\\ Latent Reasoning Chain Extraction}

\author{
    Pengze Li\textsuperscript{\rm 1, 2},
    Jiaqi Liu\textsuperscript{\rm 2, 3},
    Junchi Yu\textsuperscript{\rm 4},
    Lihao Liu\textsuperscript{\rm 2},\\
    Mingyu Ding\textsuperscript{\rm 3},
    Wanli Ouyang\textsuperscript{\rm 2, 5},
    Shixiang Tang\textsuperscript{\rm 2, 5}$^*$,
    Xi Chen\textsuperscript{\rm 1, 6}\thanks{Corresponding author}
}
\affiliations{
    \textsuperscript{\rm 1}Artificial Intelligence Innovation and Incubation Institute of Fudan University\\
    \textsuperscript{\rm 2}Shanghai Artificial Intelligence Laboratory\\
    \textsuperscript{\rm 3}UNC-Chapel Hill\\
    \textsuperscript{\rm 4}Oxford University\\
    \textsuperscript{\rm 5}The Chinese University of Hong Kong\\
    \textsuperscript{\rm 6}Shanghai Academy of AI for Science

}

\usepackage{bibentry}

\begin{document}

\maketitle

\begin{abstract}
Large language models (LLMs) are increasingly used in scientific domains. While they can produce reasoning-like content via methods such as chain-of-thought prompting, these outputs are typically unstructured and informal, obscuring whether models truly understand the fundamental reasoning paradigms that underpin scientific inference. To address this, we introduce a novel task named L\underline{a}tent \underline{R}easoning \underline{Ch}ain \underline{E}xtraction (\underline{ARCHE}), in which models must decompose complex reasoning arguments into combinations of standard reasoning paradigms in the form of a Reasoning Logic Tree (RLT). In RLT, all reasoning steps are explicitly categorized as one of three variants of Peirce’s fundamental inference modes: deduction, induction, or abduction.
To facilitate this task, we release ARCHE Bench, a new benchmark derived from 70 Nature Communications articles, including more than 1,900 references and 38,000 viewpoints. We propose two logic‑aware evaluation metrics: Entity Coverage (EC) for content completeness and Reasoning Edge Accuracy (REA) for step-by-step logical validity. Evaluations on 10 leading LLMs on ARCHE Bench reveal that models exhibit a trade-off between REA and EC, and none are yet able to extract a complete and standard reasoning chain. These findings highlight a substantial gap between the abilities of current reasoning models and the rigor required for scientific argumentation.

\end{abstract}

\begin{links}
    \link{Code}{https://github.com/Linsonng/ARCHEBenchmark/}

\end{links}

\begin{figure*}[t]
    \centering
    \includegraphics[width=0.8\linewidth]{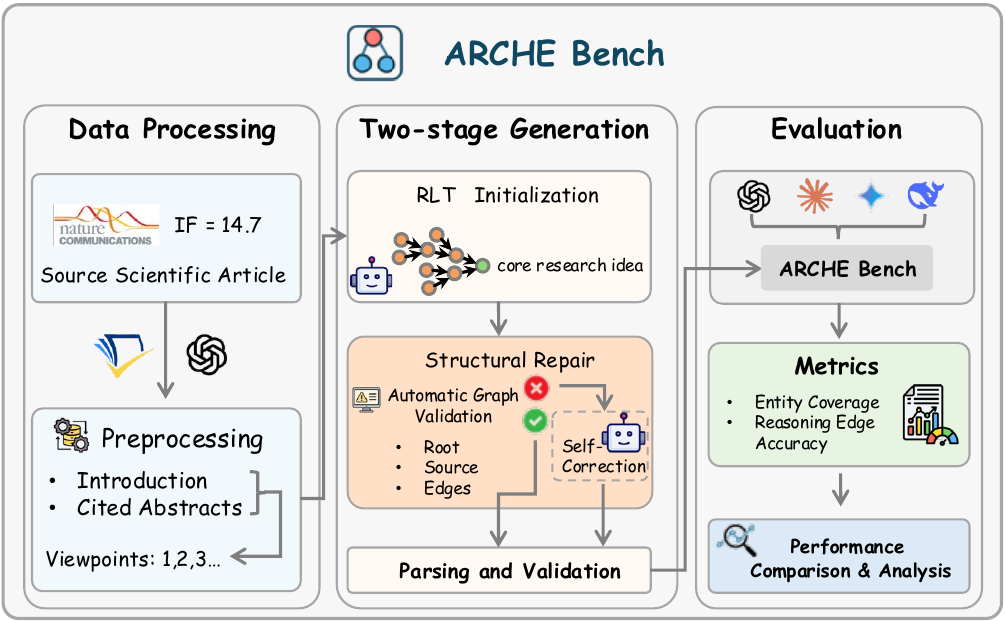}
    \caption{Overview of \textbf{\alias}.The pipeline includes three stages: Data Processing, where scientific articles are preprocessed to extract \textit{introductions}, cited abstracts, and viewpoints; RLT Generation, which constructs and repairs reasoning logic trees (RLTs) through automatic validation and self-correction; and Evaluation, where models are assessed using metrics like EC and REA for performance analysis.}
    \label{fig:overview}
\end{figure*}

\section{Introduction}

\begin{flushleft}
\textit{``All valid reasoning is either deductive, inductive, or hypo- thetic; or else it combines two or more of these characters.''}
\end{flushleft}
\begin{flushright}
\small — Charles S.~Peirce
\end{flushright}

LLMs are increasingly applied in scientific domains, from assisting literature reviews~\citep{wang2406autosurvey,zhu2025deepreview} and hypothesis generation~\citep{gottweis2025towards,xiong2024improving} to aiding in experimental design~\citep{huang2024crispr,li2025can}. Although these advances suggest the potential of LLMs to accelerate scientific discovery\cite{bai2025intern}, it is poorly understood how well these models understand and emulate human reasoning. In particular, their ability to follow and generate structured paradigm-based reasoning is uncertain, raising concerns about the trustworthiness of LLM-driven scientific workflows.

Building on Charles S. Peirce’s taxonomy, which holds that all valid reasoning is deductive, inductive, abductive, or some combination thereof~\citep{peirce1868some}, we argue that the ability to understand and appropriately apply these elementary paradigms of reasoning is essential for LLMs to perform trustworthy scientific reasoning.
Human scientists often employ a mixed inferential strategy to navigate vast bodies of evidence and competing claims in order to generate novel insights. 
Such discoveries are not isolated leaps, but chains of reasoning steps that traverse multiple paradigms. 
This process involves latent reasoning chains composed of implicit, unspoken steps that connect existing knowledge to new insights.
However, existing benchmarks fail to evaluate whether LLMs can
(i) recognize the three reasoning paradigms from complex argument,
(ii) incorporate them into coherent reasoning chains, and
(iii) ground each step in verifiable textual evidence~\cite{Yang2024Logical, hu2025survey}.

To address limitations (i) and (ii), we propose a novel task, \textbf{L\underline{a}tent \underline{R}easoning \underline{Ch}ain \underline{E}xtraction (\underline{ARCHE})}, designed to recognize the underlying reasoning behind scientific claims.
ARCHE leverages an LLM to extract fine‑grained reasoning steps from a paper’s \textit{introduction} paragraph and classify each step as deductive, inductive, or abductive reasoning. These steps are then assembled into a structured \textbf{Reasoning Logic Tree (RLT)}, where nodes correspond to individual premise or conclusion sentences, and labeled edges link each set of premise nodes to the corresponding conclusion node, indicating the associated inference type.
By (i) recognizing distinct reasoning paradigms and (ii) assembling them into an RLT, ARCHE delivers a faithful, structured representation of complex scientific arguments.

To (iii) ground the reasoning steps in real tasks, we introduce \textbf{\alias}, a benchmark derived from 70 peer-reviewed Nature Communication articles. For each paper, we provide its \textit{introduction} along with the relevant background viewpoints. We also propose two complementary metrics that capture different aspects of RLT: (a) EC, which measures proportion of key scientific entities of each paper that appear in the predicted RLT, reflecting how comprehensively the model captures the core contents; and (b) REA, which assesses the accuracy of each individual inference step in the reasoning chain, using an LLM-based judge to verify whether the conclusion logically follows from its premises given the labeled inference type. While EC ensures that no critical pieces of the core idea are omitted, REA directly evaluates logical validity at the step level.

We perform zero-shot evaluations on 10 state-of-the-art LLMs, revealing that even the best models struggle to exceed an accuracy of 50\%, indicating their limited ability to recognize and properly formalize latent reasoning chains via standard paradigms.
Furthermore, we observe a trade-off between EC and REA: models that achieve higher EC often do so at the cost of lower REA, and vice versa.

Our contributions are summarized as follows:
\begin{itemize}
    \item 
    We define a new task, Latent Reasoning Chain Extraction, to address the lack of evaluation on whether LLMs can model fundamental reasoning paradigms underlying human scientific reasoning.

    \item We design an automated pipeline for constructing \alias, a dataset of scientific articles enriched with references and extracted viewpoints. We also introduce two evaluation metrics, EC and REA, to assess the completeness and correctness of generated RLTs.

    \item We benchmark 10 state-of-the-art LLMs and reveal that models often fail to extract and formalize reasoning chains in a structurally valid way. This highlights a fundamental gap between natural language fluency and paradigm-grounded scientific reasoning, distinguishing current LLM behavior from human expert inference.

\end{itemize}

\section{Related Work}
\label{sec:related}
\subsection{Reasoning in LLMs}

Chain-of-Thought (CoT) prompting has become a foundational technique for eliciting multi-step reasoning in LLMs. Pioneered by ~\citep{wei2022cot} with few-shot examples and later shown to be effective in zero-shot settings ~\citep{kojima2022}, CoT encourages models to generate intermediate reasoning steps.
Subsequent work has enhanced its reliability through methods like self-consistency \citep{wang2023self, yuthought} and Tree-of-Thought \citep{yao2023tree}. However, CoT produces an untyped narrative, lacking the formal logical grounding needed for verifiable reasoning. This limitation has spurred a move towards logic-aware systems. One direction is neuro-symbolic, where LLMs translate natural language into formal logic for an external solver, as seen in systems like LINC \citep{Olausson2023LINC}. While formally robust, these methods can be brittle. Critically, most existing approaches study deduction, induction, and abduction in isolation, motivating our work on a unified framework.

\subsection{LLMs for Scientific Discovery}
In the scientific domain, LLMs are increasingly used to accelerate discovery, from domain-specific models that master literature like Galactica \citep{taylor2022} to systems that actively generate testable hypotheses or scientific inference\citep{nicholas2025, alkan2025, wang2025scireasoner}. Hybrid approaches combining LLMs with structured knowledge graphs have been particularly effective at generating novel and valid hypotheses \citep{tong2024}. A parallel line of work focuses on evidence synthesis, where models must aggregate information to verify scientific claims \citep{wadden2020, wan2025deepresearch}. To bring more structure to this process, the EntailmentBank benchmark requires models to construct deductive proof trees \citep{dalvi2021}. 
However, these primarily focus on textual entailment or simple deduction. There remains a critical gap in evaluating an LLM's ability to perform complete scientific reasoning—integrating inductive generalization, deductive application, and abductive explanation within a single, coherent argument grounded in evidence. \alias  addresses this gap.

\subsection{Evaluation of Scientific Reasoning}
The evaluation of LLM reasoning is a major challenge. Broad benchmarks like MMLU \citep{Hendrycks2021Measuringb} and specialized ones for math or logic \citep{Cobbe2021Training, Yu2019ReClor, wang2025physunibench} typically focus on final-answer accuracy, which can mask an unsound reasoning process. Even when reasoning types are considered, they are often evaluated in separate, targeted benchmarks \citep{Clark2020Transformers, Yang2024Language}. In the scientific context, this gap persists; existing benchmarks do not require models to explicitly identify, differentiate, and compose all three Peircean inference types within a single, coherent argument. 
Our work takes a complementary approach: rather than outsourcing logic, we challenge LLMs to produce reasoning chains with explicit, typed inference steps. By requiring models to construct a graph of deductive, inductive, and abductive moves, ARCHE-bench provides a new lens for evaluation.

\section{Methodology}
\label{sec:method}

In this section, we formalize the task of L\underline{a}tent \underline{R}easoning \underline{Ch}ain \underline{E}xtraction (\underline{ARCHE}) and describe the Reasoning-Logic Tree (RLT) designed to encode such reasoning. We then introduce \alias, a dataset constructed specifically for this task.

\subsection{Task Definition}
\label{subsec:task denition}

The core objective of this task is to interpret complex paragraphs of scientific reasoning as a combination of individual premise viewpoints and standard reasoning paradigms. 

The input of the task includes both the full \textit{introduction} section of a scientific paper and \textit{viewpoints} extracted from two sources: the \textit{introduction} itself, and the \textit{abstracts} of papers cited within that \textit{introduction}. Following the definition by Feng et al. \cite{feng2025grapheval}, a viewpoint refers to an idea, argument, or fact embedded within the research content. In our task, such viewpoints have already been extracted from the raw text using LLMs, prior to being input to any model.

The output RLT is a directed graph in DOT, a graph description language. 
A proper RLT should capture both the step-by-step inferential structure and the relevant viewpoints from the original paper or its references. Specifically, the model is expected to reconstruct how \textit{asserted facts} (stated in the \textit{introduction}), \textit{established knowledge} (from cited abstracts) and \textit{ implicit domain knowledge} (unstated yet assumed by the author) are logically connected via elementary reasoning paradigms to derive the reasoning within the paper. This demonstrates whether the model has the capacity to extract the reasoning chain behind a scientific claim and to ground each inference step into a verifiable source, fulfilling the core objective of the ARCHE task.

\subsection{Reasoning-Logic Tree (RLT)}
\label{subsec:RLT}
RLT is a structured and logic-oriented graph representation designed to make implicit reasoning steps within scientific discourse explicit, intelligible, and machine-readable. In this section, we elaborate on its nodes, edges, and graph-level constraints, as well as the pipeline used for RLT generation.

\begin{figure}[t]
    \centering
    \includegraphics[width=1\linewidth]{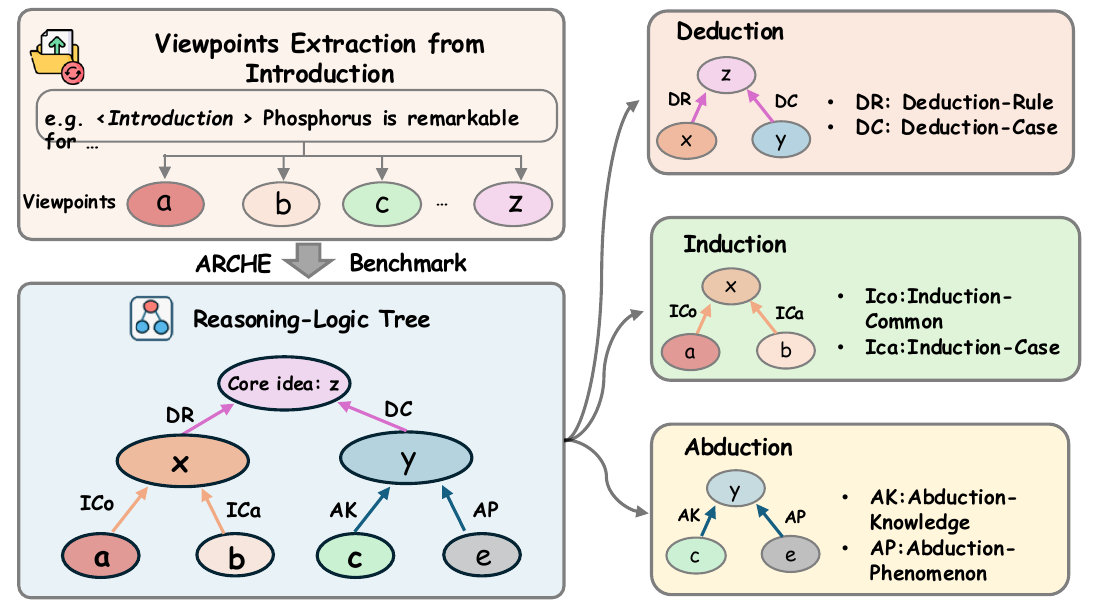}
    \caption{The Construction of Reasoning-Logic Tree. Viewpoints are first extracted from scientific text, then organized into a hierarchical reasoning structure. Each reasoning edge is annotated with an inference type: deduction, induction, or abduction, based on its logical pattern.}
    \label{fig:RLT}
\end{figure}

\paragraph{Node Definition.}
Each node in the RLT contains a viewpoint and its source. The source of a viewpoint can be (i) a sentence or a viewpoint from the \textit{Introduction} section, (ii) a viewpoint from a reference, or (iii) a sentence added by the LLM. In the third case, the added sentence often serves either as supplementary knowledge or implicit premises not explicitly stated by the author, or as an intermediate step required to bridge multi-hop reasoning. Further details of this coordinate system are provided in the appendix.

\paragraph{Edge Definition.}
The edges in an RLT indicate the reasoning paradigms underlying each inference. Each edge is directed and labeled, connecting a single premise node to a conclusion node. We define six specific edge types, which serve as fine-grained instances of Peirce's three classical paradigms of reasoning: deduction, induction, and abduction. In practice, a reasoning step should involve multiple edges, connecting at least two premise nodes to a shared conclusion node. The six edge types are formally defined below.

\begin{itemize}
\item \textbf{Deduction-Rule (DR):} A premise stating a general principle, law, or established rule.
\item \textbf{Deduction-Case (DC):} A premise that presents a specific instance or case that falls under the general rule.
\item \textbf{Induction-Common (ICo):} A premise that abstracts a common pattern or regularity across multiple cases. 
\item \textbf{Induction-Case (ICa):} A premise providing a specific observation. 
\item \textbf{Abduction-Knowledge (AK):} A premise stating existing knowledge or known mechanisms.
\item \textbf{Abduction-Phenomenon (AP):} A premise describing an observation requiring explanation.
\end{itemize}

\paragraph{Graph-Level Structural Constraints.}
To ensure that each RLT captures a coherent and focused reasoning path, we impose the following structural constraints. Each RLT is structured as a single-rooted directed acyclic graph (DAG). All nodes in the graph must converge at the central node, and disconnected or irrelevant premises are strictly prohibited. Furthermore, we enforce a standardized inference granularity by requiring that each logical step corresponds to a single reasoning paradigm, represented using one pair from the six predefined edge types. In cases of multi-hop reasoning, LLM is expected to introduce intermediate nodes that decompose the reasoning chain into multiple steps, each aligned with a specific paradigm. Edges may also cross document boundaries. For example, an edge may point from a viewpoint in a cited reference to one in the main paper, demonstrating how prior work provides premises for new research.

\paragraph{RLT Generation Pipeline.}
We design a fully automated two-stage pipeline to guide the LLM in generating a valid RLT from the source paper:

\begin{enumerate}
    \item \textbf{Stage 1: Primary Extraction.} The LLM is first guided by a manually crafted prompt to extract the latent reasoning chain and produce an initial graph in the DOT language. 
    The main ideas from Section~\ref{subsec:task denition} and Section~\ref{subsec:RLT} are included in the prompt.
    The full prompts are provided in the appendix.

    \item \textbf{Stage 2: Structural Repair.} A verifier script automatically inspects the LLM-generated RLT for structural defects, such as multiple roots, cycles, disconnected nodes, or invalid edge labels. If no defects are found, this stage is skipped. Otherwise, the same LLM is re-prompted with a targeted prompt that instructs it to correct the structural errors while preserving the original reasoning content. 

\end{enumerate}

\begin{algorithm}[tb]
\caption{Evaluation for EC and REA}
\label{alg:evaluation}
\textbf{Input}: paper text $\rm{P}$, generated RLT\\
\textbf{Output}:  $\rm{EC}$, $\rm{REA}$
\begin{algorithmic}[1]
\STATE $\text{core\_entities} \leftarrow \text{ExtractCoreEntities}(\rm{P})$ 
\STATE $\text{reasoning\_steps} \leftarrow \text{ParseReasoningSteps}(\rm{RLT})$ 

\FOR{each step $s \in \text{reasoning\_steps}$}
    \IF{$\text{ValidFormat}(s)$}
        \STATE $\text{votes} \leftarrow \text{ThreeModelVoting}(s)$ 
        \STATE $\text{validation\_results}[s] \leftarrow \text{MajorityVote}(\text{votes})$
    \ELSE
        \STATE $\text{validation\_results}[s] \leftarrow \text{format\_error}$
    \ENDIF
\ENDFOR
\STATE $\rm{REA} \leftarrow \frac{|\{s : \text{validation\_results}[s] = \text{"correct"}\}|}{|\text{reasoning\_steps}|}$
\STATE $\text{correct\_nodes} \leftarrow \{n : n \in s, \text{validation\_results}[s] = \text{"correct"}\}$
\STATE $\text{reasoning\_entities} \leftarrow \text{ExtractEntities}(\text{correct\_nodes})$ 
\STATE $\rm{EC} \leftarrow \frac{|\text{core\_entities} \cap \text{reasoning\_entities}|}{|\text{core\_entities}|}$
\STATE \textbf{return} $(\rm{EC}, \rm{REA})$
\end{algorithmic}
\end{algorithm}

\subsection{\alias}
\label{subsec:benchmark}

Building on the ARCHE task and the RLT representation detailed above, we introduce \textbf{\alias}, a benchmark designed to evaluate LLMs on real scientific texts. It consists of 70 peer-reviewed articles published in \textit{Nature Communications} in 2025. For each article, the \textit{Introduction} section is extracted, along with viewpoints obtained from the \textit{introduction} itself and from the abstracts of its cited references, using the Semantic Scholar API and GPT-4o. To support the evaluation, we propose an automated evaluation framework (Algorithm~\ref{alg:evaluation}) along with two complementary metrics.

\begin{table*}[t]
\centering

\resizebox{1\linewidth}{!}{
\begin{tabular}{>{\raggedright\arraybackslash}p{6cm}
                >{\centering\arraybackslash}p{1.5cm}
                >{\centering\arraybackslash}p{1.5cm}
                >{\centering\arraybackslash}p{1.5cm}
                >{\centering\arraybackslash}p{1.5cm}
                >{\centering\arraybackslash}p{1.5cm}
                >{\centering\arraybackslash}p{1.5cm}}
\toprule
\textbf{Model} & \multicolumn{2}{c}{\textbf{Physical Sciences}} & \multicolumn{2}{c}{\textbf{Biological Sciences}} & \multicolumn{2}{c}{\textbf{Overall}} \\
 & REA($\uparrow$) & EC($\uparrow$) & REA($\uparrow$) & EC($\uparrow$) & REA($\uparrow$) & EC($\uparrow$) \\
\midrule
Claude-Opus-4 (Thinking) & 25.5\% & 68.1\% & 22.9\% & 71.2\% & 24.2\% & \textbf{69.7\%} \\
Claude-Sonnet-4 (Thinking) & 28.6\% & 47.2\% & 29.0\% & 58.9\% & 28.8\% & 53.1\% \\
DeepSeek-R1 & 16.3\% & 31.7\% & 24.0\% & 25.6\% & 20.1\% & 28.7\% \\
Doubao-Seed-1.6 (Thinking) & 22.6\% & 55.6\% & \textbf{46.2\%} & 54.3\% & 28.2\% & 55.3\% \\
Gemini-2.5-Pro & \textbf{38.4\%} & 58.8\% & 40.5\% & 54.5\% & 39.5\% & 56.7\% \\
Gemini-2.5-Pro (Thinking) & 38.0\% & 49.4\% & 44.9\% & 59.2\% & \textbf{41.4\%} & 54.1\% \\
GPT-4o & 12.5\% & 28.7\% & 19.1\% & 19.8\% & 15.8\% & 24.3\% \\
Grok-3 & 36.0\% & 63.3\% & 30.2\% & 44.3\% & 33.1\% & 53.8\% \\
Grok-4 & 24.3\% & 55.7\% & 19.9\% & \textbf{68.2\%} & 22.2\% & 61.7\% \\
o3 & 32.3\% & \textbf{64.4\%} & 44.3\% & 50.2\% & 35.6\% & 60.5\% \\
\bottomrule
\end{tabular}
}
\caption{Performance Comparison of LLMs on ARCHE.}
\label{tab:performance_comparison}
\end{table*}

\textbf{Dataset Construction and Benchmark Statistics.} For each article in \alias, the \textit{introduction} section is extracted as the primary input for Stage 1 of RLT generation. All sentences from the \textit{introduction} and from the abstracts of cited references are parsed into viewpoints using GPT-4o. Each sentence and its corresponding viewpoints are indexed to support precise referencing during generation. The ARCHE bench comprises 70 peer-reviewed and open-access articles published in 2025, ensuring that the content is up-to-date and scientifically verified. The benchmark is balanced across two major scientific domains: Biological Sciences (35) and Physical Sciences (35). In total, the corpus contains 2,164 sentences from the articles' \textit{introductions} and 1,891 cited references, from which over 38,000 distinct viewpoints have been extracted for reasoning analysis. On average, each article's \textit{introduction} provides 30.9 sentences and 77.4 viewpoints, and is supported by 27 citations. More details are provided in Table \ref{tab:r4bench_overall}.

\textbf{Evaluation.} The algorithm~\ref{alg:evaluation} outlines the evaluation procedure for computing the EC and REA metrics. In the following, we elaborate on the key steps and underlying logic for each metric.

\textbf{Entity Coverage (EC).}  
This metric assesses the extent to which the generated RLT captures the core scientific concepts of the input article. Lines 1, 9–11 of Algorithm~\ref{alg:evaluation} describe how EC is calculated. First, \texttt{ExtractCoreEntities(P)} uses the o3 model~\citep{gpto4mini} to extract the core scientific idea from the \textit{introduction} of the article, identifying the main hypothesis or methodology. From this, scientific entities, concrete concepts, methods, and phenomena are extracted (typically 8-10). 
Coverage is calculated as the proportion of extracted entities that appear in the nodes involved in correct reasoning steps in the generated RLT.  
Entity matching is performed using case-insensitive string comparison.

\textbf{Reasoning Edge Accuracy (REA).}  
This metric measures the percentage of individual reasoning steps in the generated RLT that are logically valid. Lines 2–8 compute REA by validating each reasoning step in the generated RLT. Evaluation begins by identifying the root node and collecting all reasoning paths connected to it.
 Each step is validated for format and categorized into reasoning types. Steps with inconsistent or invalid combinations, such as pairing a deduction case with an abduction-knowledge edge, or using non-standard labels like 'deduction-knowledge', are automatically marked incorrect (line 7).  
Valid steps are evaluated using a three-model voting system (\texttt{ThreeModelVoting}, line 5), where \textit{o3}~\citep{gpto4mini}, \textit{Claude-Sonnet-4-thinking}~\citep{claude4sonnet}, and \textit{Gemini 2.5 Pro}~\citep{gemini25} independently judge whether the conclusion follows logically from the premises. A majority vote determines the final label for each step (line 6). REA is then computed as the proportion of reasoning steps labeled as correct (line 8).
We evaluated the voting system across three reasoning types using human annotations, finding that the joint model consistently outperforms individual models in alignment with human judgments, achieving an accuracy exceeding 88\%.

\begin{table}[ht]
\centering

\resizebox{1\linewidth}{!}{ 
\begin{tabular}{>{\raggedright\arraybackslash}p{7cm} r}
\toprule
\textbf{Overall} & \\
\midrule
Total Articles & 70 \\
Total Sentences & 2,164 \\
Total Viewpoints & 5,418 \\
Total Citations & 1,891 \\
Total Referenced Viewpoints & 33,321 \\
\textbf{Total Viewpoints (Combined)} & \textbf{38,739} \\
\midrule
\textbf{Average per Article} & \\
\midrule
Sentences & 30.9 \\
Viewpoints & 77.4 \\
Citations & 27.0 \\
Viewpoints per Sentence & 2.5 \\
\midrule
\textbf{Publication Year} & \\
\midrule
2025 & 70 \\
\bottomrule
\end{tabular}
}
\caption{\alias{} Overall and Average Statistics.}
\label{tab:r4bench_overall}

\end{table}

\begin{figure*}[t]
\centering
\includegraphics[width=1\textwidth]{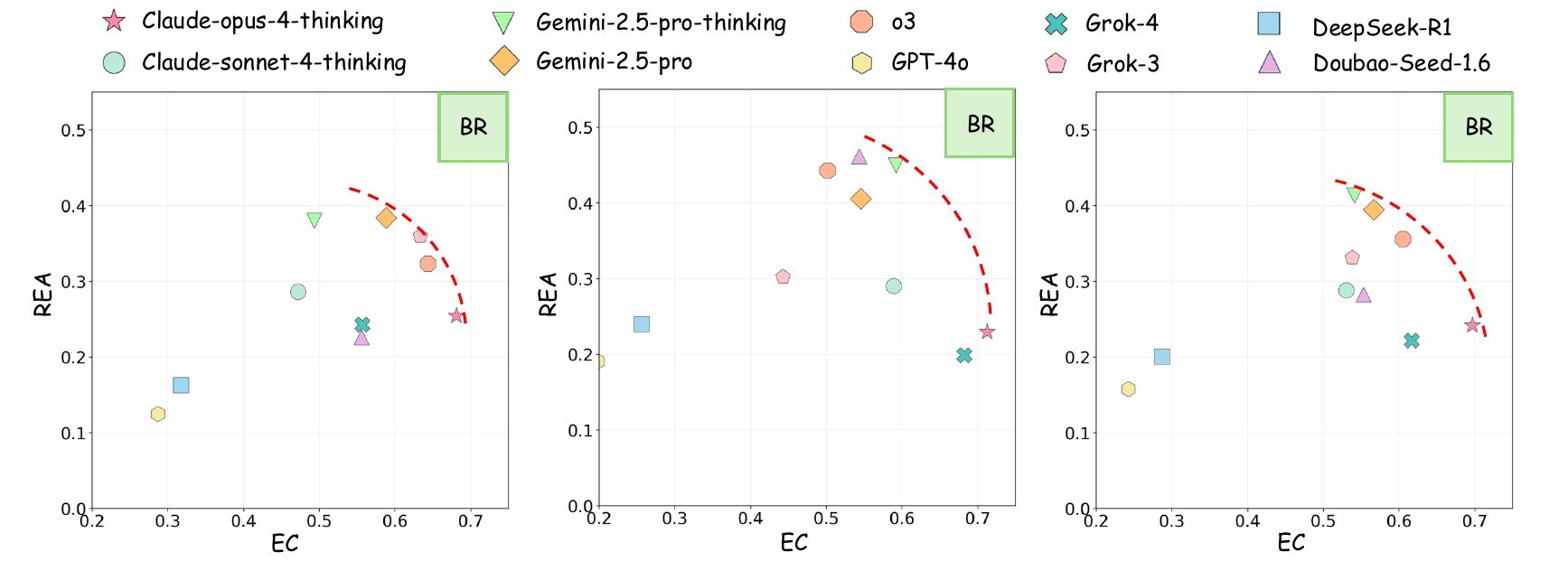}
\caption{Comparative performance of LLM models in terms of EC and REA across three domains: (Left) Physical Sciences, (Middle) Biological Sciences, and (Right) Overall. Each point represents a model’s performance. The green region (BR) indicates a preferable area with both higher coverage and accuracy. The red dashed curve denotes the trade-off frontier.}
 \label{fig:compa}
\end{figure*}

\section{Experiments}

\subsection{Models}
\label{sec:models}

We evaluated 10 leading LLMs that span six model families. These include \textit{Claude-Opus-4-thinking}~\citep{claude4sonnet}, \textit{Claude-Sonnet-4-thinking}~\citep{claude4sonnet}, \textit{GPT-4o}~\citep{openai2024gpt4ocard}, \textit{o3}~\citep{gpto4mini}, \textit{DeepSeek-R1}~\citep{DeepSeek-AI2025DeepSeekR1}, \textit{Gemini-2.5-Pro}~\citep{gemini25}, \textit{Gemini-2.5-Pro-thinking}~\citep{gemini25}, \textit{Grok-3}~\citep{grok3},\textit{ Grok-4}~\citep{grok4}, and \textit{Doubao-Seed-1.6-thinking}~\citep{doubaoseed}.  

All models are evaluated in the ARCHE task (Section~\ref{subsec:task denition}) under same conditions, using the same prompts for consistency. Temperature is fixed at 0.1 when configurable. We report performance using the EC and REA metrics described in Section~\ref{subsec:benchmark}.

\subsection{Performance Comparison and Analysis}
\label{subsec:comparison}

The overall results show that the models demonstrate moderate performance across both evaluation dimensions. 

The average EC is 51.4\%, with a wide spread from 0\% to 100\%. The median is 66.7\%, indicating that the models are able to identify most of the core scientific entities, despite high variance. REA is relatively low, with an average of 28.3\% and a median of 25\%. 
Additional results are provided in Table~\ref{tab:performance_comparison}.

As shown in Figure~\ref{fig:compa}, no model performs ideally in both EC and REA. A perfect RLT should demonstrate 100\% accuracy in tracing reasoning steps from the given viewpoints to the core idea, while simultaneously covering all key entities, as illustrated by the green region in the figure.

In terms of overall performance, models with advanced reasoning capabilities, such as \textit{Gemini-2.5-Pro-thinking}, \textit{o3}, \textit{Grok-3}, and \textit{Claude-Opus-4-thinking},  perform significantly better than comparatively less optimized models like \textit{gpt-4o}. This suggests that existing techniques, such as chain-of-thought prompting and reasoning-specific optimization strategies, indeed play a meaningful role in enhancing model reasoning. These observations also indirectly validate the utility of the ARCHE benchmark. For example, \textit{ o3} significantly outperforms \textit{GPT-4o} on both EC and REA, while \textit{Claude-Opus-4} achieves much higher EC scores than \textit{Claude-4}. On the other hand, models with similar architectures and training scales exhibit comparable behavior. For example, \textit{Gemini-2.5-Pro} and its thinking variant show only minor differences, with the latter producing a slight improvement in REA but no substantial overall gain.

However, the highest-performing models interestingly appear to align along a smooth curve (highlighted by the red dotted line), indicating that despite differences in model architecture and training objectives, there exists an apparent boundary in the joint space of accuracy and coverage. From this perspective, no single model has yet demonstrated a substantially superior ability to master and formalize reasoning chains with the fundamental paradigms. 

\subsection{Reasoning Performance by Type}
\label{subsec:bytype}

Table~\ref{tab:bytype} presents the performance of the model in three types of reasoning: abductive, deductive, and inductive. \textit{Grok-3} emerges as the model that performs the best overall, achieving the highest scores in all categories. \textit{GPT-4o} and \textit{Gemini-2.5-Pro} also show strong results, particularly in deductive reasoning. \textit{Claude-Sonnet-4} performs well on abductive tasks, with the accuracy reaching 63.7\%.

\begin{table}[ht]
\centering

\resizebox{1\linewidth}{!}{
\begin{tabular}{
  >{\raggedright\arraybackslash}p{4cm}   
  >{\centering\arraybackslash}p{1.3cm}     
  >{\centering\arraybackslash}p{1.3cm}     
  >{\centering\arraybackslash}p{1.3cm}     
}
\toprule
\textbf{Model} & \textbf{Abductive} & \textbf{Deductive} & \textbf{Inductive} \\
\midrule
Claude-Opus-4 & 58.9\% & 42.4\% & 57.1\% \\
Claude-Sonnet-4 & 62.0\% & 50.0\% & 63.7\% \\
DeepSeek-R1 & 48.8\% & 40.6\% & 59.0\% \\
Doubao-Seed-1.6 & 54.1\% & 46.9\% & 48.0\% \\
GPT-4o & 56.9\% & 63.4\% & 59.3\% \\
Gemini-2.5-Pro & 60.3\% & 59.5\% & 56.7\% \\
Gemini-2.5-Pro-thinking & 72.5\% & 56.9\% & 55.5\% \\
Grok 3 & 87.1\% & 74.0\% & 77.9\% \\
Grok 4 & 58.3\% & 36.6\% & 40.0\% \\
o3 & 57.4\% & 40.0\% & 42.2\% \\
\bottomrule
\end{tabular}
}
\caption{Reasoning Accuracy on 3 Types of Tasks.}
\label{tab:bytype}

\end{table}

In contrast, models such as \textit{DeepSeek-R1} and \textit{o3} show relatively weak performance in deductive reasoning, both scoring around 40\%. Although \textit{Grok-4} belongs to the same family as \textit{Grok-3}, its performance is notably lower in both deductive and inductive tasks. This discrepancy is likely due to \textit{Grok-4} generating more outputs without providing valid reasoning, which reduces step-level accuracy.

Note that Table~\ref{tab:bytype} only includes reasoning steps with valid output formatting. As a result, accuracy values are substantially higher than those reported in Table~\ref{tab:performance_comparison}, which also accounts for formatting errors. Despite prompt refinement and multiple rephrasings, formatting issues remain prevalent. This suggests that the current setup already approaches the expressive limits of these models. The root cause lies in their inability to grasp the structural constraints of the reasoning paradigms and to express them in syntactically valid forms.
This also explains why \textit{GPT-4o} does not appear significantly weaker than other models in Table~\ref{tab:bytype}. A large portion of its output contains structural violations, such as mixing incompatible reasoning types, which are filtered out during the format validation. This further demonstrates the robustness of our benchmark, which reliably distinguishes models by their ability to generate structurally valid and correct reasoning, rather than surface-level fluency.

\begin{table}[h]
\centering

\resizebox{1\linewidth}{!}{
\begin{tabular}{
  >{\raggedright\arraybackslash}p{4cm} 
  >{\centering\arraybackslash}p{1.5cm}      
  >{\centering\arraybackslash}p{1.5cm}      
}
\toprule
\textbf{Model} & \textbf{ATS} & \textbf{AES} \\
\midrule
Gemini-2.5-Pro & 12.4 & 5.8 \\
Gemini-2.5-Pro-thinking & 13.2 & 5.3 \\
o3 & 11.7 & 4.9 \\
Grok-4 & 20.1 & 4.9 \\
Grok-3 & 11.0 & 4.0 \\
Claude-Opus-4 & 11.0 & 3.3 \\
Doubao-Seed-1.6 & 8.6 & 2.5 \\
Claude-Sonnet-4 & 8.1 & 2.2 \\
DeepSeek-R1 & 8.9 & 1.9 \\
GPT-4o & 9.2 & 1.2 \\
\bottomrule
\end{tabular}
}

\caption{Performance Comparison of Language Models by Step Metrics. ATS: Average Total Steps; AES: Average Effective Steps.}
\label{tab:atsaes}

\end{table}

\subsection{Analysis of Step Efficiency}
\label{subsec:step-efficiency}

Table~\ref{tab:atsaes} compares the extraction behavior of reasoning chains of different language models, using Average Total Steps (ATS) and Average Effective Steps (AES) as evaluation metrics. ATS reflects the average number of reasoning steps generated per RLT, while AES measures the subset of steps that are logically valid and relevant to the final answer.

The results show that Grok-4 has the highest ATS (20.1), indicating a tendency to produce longer reasoning chains. However, its AES (4.9) does not increase proportionally, resulting in a relatively low proportion of effective steps. In contrast, o3 achieves a similar AES (4.9) with a much shorter ATS (11.7). A similar pattern is observed in Gemini-2.5-Pro-thinking, which generates slightly more steps (ATS 13.2) than the base Gemini-2.5-Pro (ATS 12.4), but with a marginally lower AES (5.3 vs. 5.8). Models such as Claude-Sonnet-4 and DeepSeek-R1 exhibit lower ATS values (around 8–9), with correspondingly low AES scores. In particular, GPT-4o produces the lowest AES (1.2) among all models, suggesting that it typically contributes only a single effective reasoning step per instance.

These findings indicate that longer reasoning chains do not necessarily correlate with higher reasoning quality. Some models tend to produce verbose or redundant outputs without improving step utility. Furthermore, even the best-performing model extracts fewer than six valid inferences on average from \textit{introductions} containing more than 30 sentences, despite also having access to supporting viewpoints from references. This highlights a substantial gap between the performance of the current model and the goal of robust paradigm-aligned scientific reasoning.

\section{Discussion}
\label{sec:discussion}

\subsection{Key Findings and Insights}

Our experiments reveal a fundamental limitation in the ability of current LLMs to explain scientific reasoning. Although many leading large reasoning models are capable of generating fluent, reasoning-like natural language outputs, they still lack a genuine grasp of reasoning itself. Even when presented with peer-reviewed scientific texts that contain verified and logically coherent points of view, these models often fail to identify the underlying logic of reasoning behind scientific claims and evidence.

This deficiency poses a major risk in the use of LLMs for scientific discovery. Scientific discovery requires a transparent and grounded thought process for rigorous verification and supervision. To this end, we advocate for the explicit incorporation of reasoning-paradigm-aligned data during model pre-training and instruction. Alternatively, reasoning supervision objectives could be augmented with reward signals grounded in formal paradigms to promote structurally valid and interpretable inference.

This motivation underlies the name of our task, \textbf{ARCHE}. The term \textit{ arch}, originating from ancient Greek philosophy and later formalized by Aristotle, denotes 'principle', 'source' or 'cause'. It reflects our belief that an effective reasoning benchmark must go beyond surface-level correctness and instead guide the development of models capable of verifiable, paradigm-based reasoning in complex scientific domains. We hope that the ARCHE Bench will serve as a foundation for future work toward models that not only speak in the language of reasoning but also reason in its true form.

\subsection{Limitations and Future Work}

Despite the insights obtained, our study has several limitations, which we aim to address in future work:

\begin{enumerate}
  \item \textbf{Limited corpus scale.} The current benchmark includes only 70 research articles. Each ARCHE evaluation requires the model to process a full \textit{introduction} along with all cited abstracts, resulting in high token usage, approximately \$4 per paper across all 10 models. Although this constrains the size of the dataset, previous work~\citep{press2024citeme} suggests that small, carefully curated corpora can still meaningfully reveal model weaknesses. To improve both scale and diversity, we plan to expand the benchmark to include multidisciplinary content such as chemistry and artificial intelligence to assess cross-domain generalization.

  \item \textbf{Partial research context.} The current benchmark focuses solely on the \emph{\textit{Introduction}} section, as it typically presents the central hypothesis or contribution. However, excluding \emph{Methods} and \emph{Results} may underestimate an LLM’s ability to reason across the full scientific workflow. Reasoning based on experimental findings and iterative hypothesis formation is also essential. We will extend ARCHE to include these sections, allowing end-to-end evaluation of hypothesis generation, evidence collection, and conclusion formation.
\end{enumerate}

\section{Conclusion}
\label{sec:conclusion}

We present a novel task, ARCHE, along with an automated benchmark designed to evaluate LLMs on scientific reasoning. Through evaluation on \alias, we demonstrate that state-of-the-art models, including those optimized for reasoning, still struggle to reliably identify and organize reasoning chains in scientific texts. Our multi-perspective analysis reveals that fluent natural language reasoning does not imply an internalized understanding of reasoning paradigms. These findings highlight the need for paradigm-guided reasoning supervision, whether through data, reward design, or training framework. We hope that ARCHE will serve as a foundation for future research toward models that truly learn and apply reasoning.

\section*{Acknowledgement}
\label{sec:Acknowledgement}

This work is supported by Shanghai Artificial Intelligence Laboratory. This work was done during Pengze Li's internship at Shanghai Artificial Intelligence Laboratory. The computations in this research were performed using the CFFF platform of Fudan University

\bibliography{aaai2026}

\appendix

\label{appendix}

\newpage..
\newpage
\section{Benchmark Construction and Quality Assurance}
\label{app:construction}

The construction of our benchmark and its associated evaluation pipeline was guided by four core principles: \textbf{structural rigor}, \textbf{content traceability}, \textbf{logical validity}, and \textbf{reproducibility}. To uphold these principles, we implemented a multi-faceted quality assurance framework, encompassing data representation, a multi-layered validation process, and performance optimizations. This section provides further details on these technical features.

\paragraph{High-Precision Data Traceability.}
A cornerstone of our benchmark is a high-precision, three-digit coordinate system, \texttt{(x,y,z)}, that ensures every piece of information is unambiguously traceable to its origin. This system allows every node in a Reasoning-Logic Tree (RLT) to be mapped to either a whole sentence from the main text \texttt{(x,0,0)}, a specific viewpoint within that sentence \texttt{(x,y,0)}, or a viewpoint from a cited reference \texttt{(x,y,z)}. This granular approach is critical for our grounded evaluation methodology. The entire pipeline is designed to be backward-compatible, capable of parsing older two-digit formats and mapping them into the new system. We prioritize this original source content in all evaluation steps; any transcribed text on node labels serves only as a fallback.

\paragraph{A Multi-Layered Validation Framework.}
Our evaluation framework is multi-layered, first ensuring the structural integrity of a generated graph before proceeding to more complex semantic and logical validation. This tiered approach enhances the robustness and efficiency of the assessment.

At the structural layer, every generated graph undergoes a series of automated checks before any content analysis. These include verifying the existence of a single root node (the argument's final conclusion), ensuring there are no isolated nodes or subgraphs, and, most critically, enforcing the strict Peircean edge pairing rules for each inference type. Any graph failing these structural checks is either sent back for automated correction or its malformed steps are directly penalized in the final score.

At the content and logical layer, we apply our two primary metrics, EC and REA, with additional quality controls. For Entity Coverage (EC), the gold-standard entity set is extracted from the paper's core idea by a powerful LLM (o3) to establish an objective reference. The coverage calculation itself is restricted to the graph's main connected component to focus only on the primary line of argument. For Reasoning Edge Accuracy (REA), we employ a robust multi-agent validation framework to mitigate the biases of any single model. Each structurally valid inference step is judged by a panel of three distinct models (o3, GPT-4o, and Gemini). The final verdict is based on a majority vote, with a tie-breaking rule that favors a `correct` classification to avoid excessive penalization in ambiguous cases. This entire process is designed to be fault-tolerant, where the failure of a single model's API call does not disrupt the overall evaluation.

\paragraph{Performance and Reproducibility.}
To ensure this rigorous evaluation process is computationally feasible, we implemented several key optimizations. The validation of hundreds of reasoning steps across three models is fully parallelized, with each (step, model) pair treated as an independent task. The tiered filtering approach also significantly improves efficiency by preventing costly LLM API calls for structurally flawed inferences.

For reproducibility, the entire pipeline is designed for consistency. The same graph parsing libraries (\texttt{pydot}), content retrieval logic, and entity extraction algorithms are used across all stages. All intermediate results, from prompts to model responses, are cached to disk, facilitating debugging and ensuring that every evaluation run can be precisely replicated.

\section{Reasoning-Logic Tree Generation.}
\label{app:RLT_generation}

The generation of each Reasoning-Logic Tree (RLT) is performed by a large language model (LLM) through a structured, multi-step process. This process is designed to guide the LLM in producing graphs that are not only content-rich but also structurally sound and logically coherent according to Peircean inference principles.

\paragraph{Prompting with Granular Source Information.}
The initial prompt provided to the LLM is structured to supply all necessary context with high-precision source attribution. For each sentence in a paper's introduction, the prompt includes:
\begin{itemize}
    \item The original sentence text.
    \item A pre-extracted list of fine-grained \textit{viewpoints} contained within that sentence.
    \item A list of viewpoints extracted from the abstracts of all papers cited by that sentence.
\end{itemize}

Each piece of information is associated with a unique three-digit coordinate, \texttt{(x,y,z)}, allowing the model to trace every statement back to its origin. The LLM is instructed to generate nodes in a strict DOT graph format, where each node's label must begin with its source coordinate: $(x, 0, 0)$ for the $x$-th sentence itself; $(x, y, 0)$ for the $y$-th viewpoint in the $x$-th sentence; $(x, y, z)$ for the $z$-th viewpoint from the $y$‑th reference cited in the $x$‑th sentence; and $(0, 0, 0)$ for any additional sentence.

\paragraph{Two-Step Generation with Automated Verification and Correction.}
Instead of a single one-shot generation, we employ a robust two-step procedure to ensure the quality and validity of the final RLT.

\begin{description}
    \item[Step 1: Initial Generation.] The LLM first receives the comprehensive prompt and generates an initial RLT. This output is then passed to an automated validator.
    
    \item[Step 2: Automated Validation and Iterative Refinement.] The validator programmatically checks the generated graph against a set of strict structural and logical criteria:
    \begin{itemize}
        \item \textbf{Structural Integrity:} The graph must have a single root node (i.e., a final conclusion with no outgoing edges) and must not contain any isolated, disconnected nodes.
        \item \textbf{Edge Type Standardization:} All edge types must conform to one of the six predefined Peircean inference types.
        \item \textbf{Strict Edge Pairing:} Every conclusion node in the graph must be supported by a valid and complete set of premises. The validator enforces the pairing rules for each inference type: a \textit{deduction} requires exactly one `deduction-rule` and one `deduction-case` edge; an \textit{abduction} requires one `abduction-phenomenon` and one `abduction-knowledge` edge; and an \textit{induction} requires one `induction-common` and at least one `induction-case` edge.
    \end{itemize}
    If the initial graph fails any of these checks, the system automatically triggers a correction loop. It generates a new prompt that includes the original request along with a description of the detected errors, asking the LLM to regenerate the graph while fixing the specific issues. This iterative refinement step significantly improves the logical and structural quality of the output.
\end{description}

The final, validated graph is the output of this pipeline and serves as the input for the downstream evaluation metrics described in Section \ref{app:evaluation}. The entire generation process, including all intermediate prompts and responses, is logged for full reproducibility.

\paragraph{Generation Statistics Across Models.}
  We evaluated ten state-of-the-art LLMs on a dataset of 70 scientific papers to assess their ability to generate valid RLTs through our two-step pipeline. Table~\ref{tab:generation_stats} reports the Stage 2 trigger rate for each model, defined as the proportion of papers that required automated correction in Step 2 out of the total 70 papers. A \textit{lower} trigger rate indicates superior performance, as it means the model more frequently produced structurally valid and logically coherent reasoning graphs in the initial generation (Step 1) without needing iterative refinement. 

  \begin{table}[h]
  \centering
  \caption{Stage 2 trigger rates across models. The trigger rate represents the proportion of papers requiring automated correction
  (Step 2). Lower rates indicate better initial generation quality.}
  \label{tab:generation_stats}
  \begin{tabular}{lc}
  \toprule
  \textbf{Model} & \textbf{Stage 2 Trigger Rate} \\
  \midrule
  Claude-Opus-4 & 82.9\% \\
  Claude-Sonnet-4 & 58.6\% \\
  DeepSeek-R1 & 52.9\% \\
  Doubao-Seed-1.6 & 80.0\% \\
  GPT-4o & 61.4\% \\
  Gemini-2.5-Pro & 72.9\% \\
  Gemini-2.5-Pro-thinking & 74.3\% \\
  Grok 3 & 64.3\% \\
  Grok 4 & 55.7\% \\
  o3 & 67.1\% \\
  \bottomrule
  \end{tabular}
  \end{table}

\section{Evaluation Methodology}
\label{app:evaluation}

To quantitatively assess the quality of model-generated Reasoning-Logic Trees (RLTs), we propose two complementary, automated metrics: Entity Coverage (EC) to measure content completeness, and Reasoning Edge Accuracy (REA) to evaluate step-by-step logical validity. Here are more details about this metrics:

\subsection{Metric 1: Entity Coverage (EC)}
\label{subsec:ec_metric}

The EC metric evaluates how comprehensively the generated RLT captures the essential concepts of the source article's main argument. The calculation process involves three main stages.

\paragraph{Gold-Standard Core Concept Generation.}
To create an objective reference, we first establish a gold-standard set of essential concepts. This is a two-step process performed by a powerful LLM (o3):
\begin{enumerate}
    \item \textbf{Core Idea Extraction:} The model reads the source article's introduction to synthesize its single, core research idea.
    \item \textbf{Core Entity Extraction:} The model then extracts key scientific entities exclusively from this core idea. The prompt is specifically refined to prohibit entities not explicitly mentioned, ensuring high fidelity. This list serves as the ground truth of concepts that a complete RLT should cover.
\end{enumerate}

\paragraph{Entity Extraction from the Predicted Graph.}
Next, we extract entities from the model-generated graph in a robust and grounded manner.
\begin{enumerate}
    \item \textbf{Graph Parsing and Source Linking:} We parse the DOT graph and use the high-precision three-digit coordinate $(x, y, z)$ embedded in each node to link it to its original source content (e.g., a specific sentence, a viewpoint within a sentence, or a viewpoint from a cited reference).
    
    \item \textbf{Connected Component Analysis:} To focus on the main line of argument, we perform a Breadth-First Search (BFS) starting from the graph's root nodes (those with no incoming edges). This isolates the primary connected component, filtering out any disconnected or spurious nodes.
    
    \item \textbf{Grounded Entity Extraction:} For each node within this connected component, we retrieve its original source text via its coordinate. \textit{Crucially, entities are extracted from this original content, not from the node's potentially paraphrased label}, ensuring strict grounding.
\end{enumerate}

\paragraph{Calculation}
The final EC score is the percentage of gold-standard core entities found in the set of entities extracted from the graph, using a flexible containment criterion for matching.
\begin{equation}
\label{eq:ec}
\text{EC} = \frac{|\text{Covered Core Entities}|}{|\text{Total Core Entities}|} \times 100\%
\end{equation}

\subsection{Metric 2: Reasoning Edge Accuracy (REA)}
\label{subsec:rea_metric}

The REA metric assesses the logical validity of each individual inference step in the RLT using a multi-tiered methodology.

\paragraph{Structural Filtering.}
The process begins by identifying all potential inference steps (a set of premise edges converging on a conclusion node). We first categorize each step based on its structural validity:
\begin{itemize}
    \item \textbf{Format Errors:} Steps are immediately classified as errors if their premise edges do not conform to the strict Peircean pairing rules described in Section~\ref{app:RLT_generation}.
    \item \textbf{Structurally Valid Steps:} Steps with a valid logical pattern proceed to the next stage for semantic evaluation.
\end{itemize}
This initial stage efficiently removes malformed inferences without requiring expensive LLM calls.

\paragraph{LLM-based Semantic Validation.}
Each structurally valid step is then subjected to a rigorous semantic validation by a panel of three distinct, powerful LLMs acting as independent judges. For each step, a prompt is generated containing the grounded source content of the premises and the conclusion. The LLM panel's task is to determine if the conclusion logically follows from the premises.

To ensure robustness, the final verdict for each step is determined by a majority vote from the three models. In the event of a tie, the step is classified as `correct`.

\paragraph{LLM Accuracy Calibration.}
To evaluate the actual correctness of the triple-model voting system, we manually sampled a representative subset of voting instances. The samples were evenly drawn from the three types of reasoning (deduction, induction, and abduction), and stratified into four types of decision: unanimously correct, unanimously incorrect, majority correct, and majority incorrect. Each case was independently annotated by human evaluators to establish ground truth labels, as visualized in the accompanying heat map.
These annotations allow us to quantify the agreement of each model with human judgment, which we treat as a proxy for the accuracy of the true reasoning. We report both the agreement between individual models and human annotations as well as their alignment with the ensemble (triple-model) decision. Our results indicate that the joint model decision consistently outperforms any single model in terms of agreement with human annotations, suggesting that ensemble-based voting can mitigate model-specific errors and more closely approximate human reasoning standards.

\begin{figure}[t]
    \centering
    \includegraphics[width=1\linewidth]{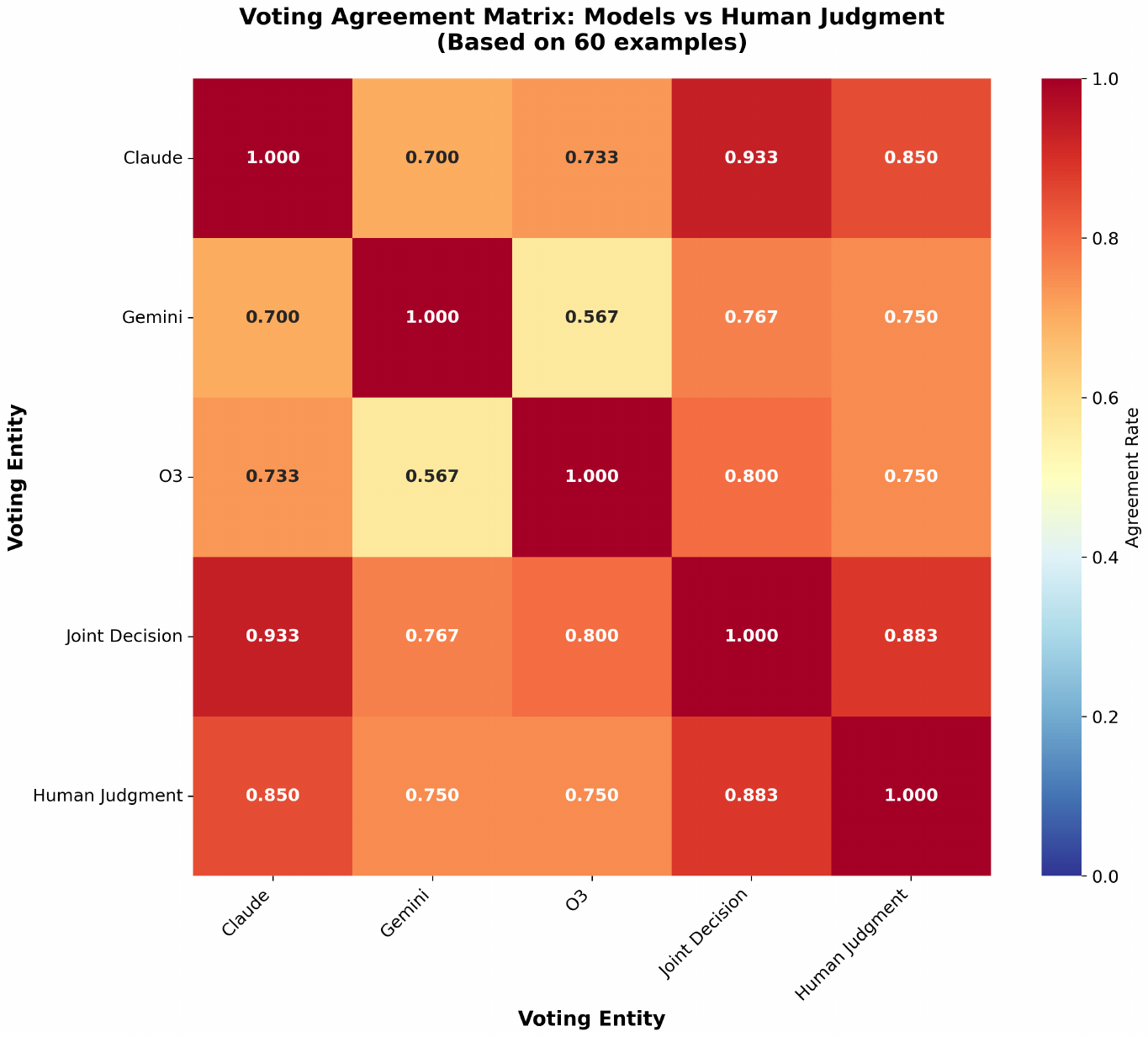}
    \caption{LLM Accuracy Calibration.}
    \label{fig:performance}
\end{figure}

\paragraph{Calculation.}
After validation, every attempted inference step is categorized as either \texttt{correct}, \texttt{wrong} (failed the LLM vote), or a \texttt{format\_error}. The REA score is the ratio of correctly validated steps to the total number of steps attempted, meaning both semantic and structural errors penalize the final score.
\begin{equation}
\label{eq:rea}
\text{REA} = \frac{|\text{Correct Steps}|}{|\text{Total Attempted Steps}|} \times 100\%
\end{equation}

\section{Case Examples and Prompts for Benchmark Construction}
\label{app:sentence_viewpoint_examples}

The following examples illustrate the extracted viewpoints and associated references from input sentences in our dataset. For brevity, some references have been truncated.


\begin{tcolorbox}[colback=gray!5!white, colframe=gray!80!black, title=Sentence 1]
\textbf{Sentence:} \\
Furthermore, seawater is a natural carbon sink of net $\sim$0.4 giga-ton CO\textsubscript{2} per year via the flux exchange between the seawater and atmosphere, potentially supporting trillion-ton-scale CO\textsubscript{2} capture, utilization, and sequestration via engineering solutions [3], [4].

\textbf{Extracted Viewpoints:}
\begin{itemize}
    \item Seawater is a natural carbon sink of net approximately 0.4 giga-ton CO\textsubscript{2} per year via the flux exchange between the seawater and atmosphere.
    \item Seawater potentially supports trillion-ton-scale CO\textsubscript{2} capture, utilization, and sequestration via engineering solutions.
\end{itemize}

\textbf{Reference Highlights (from [3]):}
\begin{itemize}
    \item Electrochemical CO\textsubscript{2} capture technologies are gaining attention for their flexibility.
    \item These methods can address decentralized emissions, such as those from the ocean and atmosphere.
    \item A pH-swing can be applied electrochemically through various techniques such as bipolar membrane electrodialysis or reversible redox reactions.
    \item (20 more entries omitted for brevity.)
\end{itemize}
\end{tcolorbox}


\begin{tcolorbox}[colback=gray!5!white, colframe=gray!80!black, title=Sentence 2]
\textbf{Sentence:} \\
In particular, the conversion of dissolved carbon in seawater into carbon-based fuels or chemical products can be entirely powered by sunlight using photoelectrochemical (PEC) devices.

\textbf{Extracted Viewpoint:}
\begin{itemize}
    \item The conversion of dissolved carbon in seawater into carbon-based fuels or chemical products can be entirely powered by sunlight using photoelectrochemical devices.
\end{itemize}
\end{tcolorbox}

\begin{tcolorbox}[colback=gray!5!white, colframe=gray!80!black, title=Sentence 3]
\textbf{Sentence:} \\
MOFs are cage-like structures containing metal nodes and organic ligands connected in infinite arrays [5].

\textbf{Extracted Viewpoints:}
\begin{itemize}
    \item MOFs are cage-like structures.
    \item MOFs contain metal nodes.
    \item MOFs contain organic ligands.
    \item MOFs are connected in infinite arrays.
\end{itemize}

\textbf{Reference Highlights (from [5]):}
\begin{itemize}
    \item MOFs are recognized as a fascinating class of materials with scientific and engineering value.
    \item MOFs possess high porosity and well-defined atomic-level structures.
    \item The microporous nature of MOFs gives them size-selective capabilities and high surface area.
    \item There is growing interest in designing hierarchically porous MOF-based materials for broader applications.
\end{itemize}
\end{tcolorbox}

\begin{figure*}[t]
    \centering
    \fcolorbox{black}{lightgray!20}{\parbox{0.95\textwidth}{
    \subsection{Prompt for ARCHE}
    
    Charles S. Peirce, a member of the National Academy of Sciences of the United States, pointed out that all valid reasoning is either deductive, inductive, or hypothetic; or else it combines two or more of these characters. Now, I have an introduction section of a scientific article. Please extract its core scientific research proposal or idea, and use the above three types of reasoning to show the process of reasoning to get the idea. In the paper fragment I provided, the content part is the complete introduction paragraph, the sentence part is the original sentence of each paper, and the reference part is the viewpoint extracted from the abstract of the reference cited by the original sentence.
    
    Please build a complete logical reasoning chain based on the content and viewpoints of the paper I provided. Please use Graphviz DOT syntax to output the entire graph and present it in the form of code blocks. Please strictly abide by the following requirements:
    
    Overall goal: Extract the logical structure behind the scientific research idea from the original text. Its structure can be a tree structure diagram, which describes the process of reasoning from various raw information to get the scientific research idea. Specifically, you will build a single-rooted reasoning tree, and the following is the specific definition.
    
    \begin{enumerate}
        \item The node is divided into two parts:
            \begin{enumerate}
                \item Source (three ints, X, Y, Z), choose one of the following four situations:
                    \begin{itemize}
                        \item Original sentence (X: sentence idx, Y: 0, Z: 0)
                        \item Original viewpoint (X: sentence idx, Y: viewpoint number, Z: 0)
                        \item Reference opinion (X: sentence idx, Y: reference number, Z: opinion number within that reference)
                        \item Implicit information for reasoning (X: 0, Y: 0, Z: 0)
                    \end{itemize}
                    
                \item Transcription (a string), considering that the original paper may not directly express the reasoning, you need to transcribe it into a sentence with a more obvious reasoning format. For example, at the starting node of the deduction-rule edge, the transcription part should include a deductive reasoning rule: If xxxx, then yyyy. At the starting node of the deduction-case edge, it should be transcribed: Currently there is xxxx. At the common end point of the deduction-rule edge and the deduction-case edge, it should be transcribed: deduction-reasoning: Currently there is xxxx, and if xxxx, then yyyy, so yyyy. The above three instructions are based on a deductive reasoning, and are required to strictly comply with the reasoning logic (i.e., the deduction-case situation should be exactly the part assumed in the deduction-rule).
            \end{enumerate}
            
        \item The edge type can only be one of the following 6 types:
            \begin{itemize}
                \item deduction-rule
                \item deduction-case
                \item abduction-phenomenon
                \item abduction-knowledge
                \item induction-case
                \item induction-common
            \end{itemize}
            
        \item CRITICAL CONSTRAINT - Edge Pairing Requirements: Every reasoning conclusion must be reached by exactly two edges of specific paired types pointing to the same target node. The valid pairs are:
            \begin{itemize}
                \item For deductive reasoning: One "deduction-rule" edge and one "deduction-case" edge must both point to the same target node
                \item For abductive reasoning: One "abduction-phenomenon" edge and one "abduction-knowledge" edge must both point to the same target node  
                \item For inductive reasoning: One "induction-case" edge and one "induction-common" edge must both point to the same target node
            \end{itemize}

    \end{enumerate}
    }}
    \caption{Prompt for ARCHE (Part 1)}
    \label{fig:scientific_reasoning_prompt_part1}
\end{figure*}

\begin{figure*}[t]
    \centering
    \fcolorbox{black}{lightgray!20}{\parbox{0.95\textwidth}{
    \subsection{Prompt for ARCHE (continued)}

    \begin{enumerate}
        \setcounter{enumi}{3}
        \item Other constraints:
        \begin{itemize}
            \item If there is multi-hop reasoning in the reasoning chain (or a single reasoning/induction/deduction is not enough to explain clearly, such compound reasoning is common in scientific literature.), please introduce intermediate nodes (as implicit information) to break down the logical path into multiple clear reasoning steps. The intermediate nodes must also be written with complete sentences and source annotations. The more detailed the reasoning and the more nodes there are, the higher the score will be.
            \item Domain consensus can be used as implicit information, but it cannot be written directly in the form of a conclusion, and must be written as callable background knowledge.
            \item If a node is both the conclusion node of a certain reasoning (this is very common in scientific reasoning) and the argument node of the next reasoning, it can contain multiple transcribed sentences. Indicate according to different edges.
            \item All nodes must serve the logical backbone, that is, all reasoning will eventually be reasoned to the same final node, which represents the determination of the scientific research idea. Therefore, there is only one root node with only input edges and no output edges; in addition, there must be no nodes that are abandoned after reasoning, and no isolated nodes that cannot be connected.
            \item The end point of the reasoning chain is to propose a scientific research idea or method, and it does not need to extend to the discussion of the results.
            \item Please ensure HIGH coverage of the original sentences (aim for at least 70
        \end{itemize}

        \item PAPER CONTENT:
        \begin{itemize}
            \item {data["introduction"]["content"]}
        \end{itemize}
        
        \item EXTRACTED SENTENCES AND VIEWPOINTS:
        \begin{itemize}
            \item {\textit{sentences info}}
        \end{itemize}
    \end{enumerate}
    }}
    \caption{Prompt for Scientific Reasoning Path Extraction (Part 2)}
    \label{fig:scientific_reasoning_prompt_part2}
\end{figure*}

\begin{figure*}[t]
    \centering
    \fcolorbox{black}{lightgray!20}{
        \begin{minipage}{0.9\textwidth}
        \small
            
            \textbf{The DOT graph has structural issues that need to be fixed. Please fix ALL structural problems while preserving content and maintaining the quality-coverage balance.}

            \textbf{DETECTED ISSUES:}
            \begin{itemize}[nosep, leftmargin=*]
                \item Single root node missing
                \item Isolated nodes found
                \item Incorrect node formatting
            \end{itemize}

            \textbf{ORIGINAL DOT GRAPH (with issues):}
first response

            \textbf{CRITICAL:} You must fix ALL issues while preserving reasoning content. You must also strictly follow ALL original format requirements below:

            \textbf{COMPLETE FORMAT REQUIREMENTS (MUST BE FOLLOWED):}
            \begin{enumerate}[nosep, leftmargin=*]
                \item \textbf{NODE FORMAT REQUIREMENTS (EXACT COMPLIANCE):}
                \begin{itemize}[nosep, leftmargin=*]
                    \item Each node MUST follow this format:
                    \texttt{node\_id [label="(source\_x,source\_y,source\_z) transcription\_content"];}
                    \item Source formats:
                    \begin{itemize}[nosep, leftmargin=*]
                        \item Original sentence: \texttt{(sentence\_idx,0,0)}
                        \item Original viewpoint: \texttt{(sentence\_idx,viewpoint\_number,0)}
                        \item Reference viewpoint: \texttt{(sentence\_idx,reference\_number,opinion\_number)}
                        \item Implicit knowledge: \texttt{(0,0,0)}
                    \end{itemize}
                    \item Transcription requirements: ``If [condition], then [consequence]''; ``Currently [specific situation]''; etc.
                \end{itemize}

                \item \textbf{EDGE TYPES (ONLY THESE 6 ALLOWED):}
                \begin{itemize}[nosep, leftmargin=*]
                    \item deduction-rule, deduction-case, abduction-phenomenon, abduction-knowledge, induction-case, induction-common
                \end{itemize}

                \item \textbf{EDGE PAIRING (ABSOLUTE REQUIREMENT):}
                \begin{itemize}[nosep, leftmargin=*]
                    \item Deductive: ``deduction-rule'' + ``deduction-case'' $\rightarrow$ same target
                    \item Abductive: ``abduction-phenomenon'' + ``abduction-knowledge'' $\rightarrow$ same target
                    \item Inductive: ``induction-case'' + ``induction-common'' $\rightarrow$ same target
                \end{itemize}

                \item \textbf{STRUCTURE REQUIREMENTS:} Exactly ONE root node; No isolated nodes.
                \item \textbf{REASONING TYPE CLARIFICATIONS:} INDUCTIVE: cases $\rightarrow$ rule; DEDUCTIVE: rule + case $\rightarrow$ conclusion; ABDUCTIVE: phenomenon + knowledge $\rightarrow$ hypothesis.
            \end{enumerate}

            \textbf{SPECIFIC FIXES REQUIRED:}
            \begin{enumerate}[nosep, leftmargin=*]
                \item SINGLE ROOT NODE: Ensure exactly one final node with no outgoing edges.
                \item NO ISOLATED NODES: Connect all nodes to the main reasoning chain.
                \item PROPER EDGE PAIRING: Each reasoning conclusion needs exactly 2 paired edges.
                \item NODE FORMAT: Fix node labels to use (X,Y,Z) source format.
                \item SINGLE SOURCE PER NODE: Each node must have exactly one source.
                \item STANDARD EDGE TYPES ONLY: Use only the 6 allowed types.
                \item PRESERVE CONTENT: Keep all reasoning transcriptions and source information.
            \end{enumerate}

            \textbf{YOUR TASK:}
            \begin{itemize}[nosep, leftmargin=*]
                \item Fix all structural issues while keeping the scientific reasoning content.
                \item Ensure every reasoning step follows the proper paired-edge pattern.
                \item Return ONLY the corrected DOT format graph, wrapped in \texttt{\`{}\`{}\`{}dot} code blocks.
            \end{itemize}

            \textbf{PAPER CONTENT FOR REFERENCE:}
            input data["introduction"]["content"]

            \textbf{EXTRACTED SENTENCES FOR REFERENCE:}
            sentences info
        \end{minipage}%
}
    \caption{The complete, compact prompt for automated structural correction. Spacing and list formats have been adjusted to fit on a single page.}
    \label{fig:structure_check_prompt_compact}
\end{figure*}

\begin{figure*}[t]
    \centering
    \fcolorbox{black}{lightgray!20}{%
        \begin{minipage}{0.9\textwidth}
        \small 

            \textbf{DOT Graph Content:}

            \begin{enumerate}[nosep, leftmargin=*]
                \item \textbf{Deductive reasoning for the need for new carbon capture technologies:}
                \begin{itemize}[nosep, leftmargin=*]
                    \item "1" [label="(1, 0, 0) Developing carbon capture, utilization, and storage technologies is crucial for managing anthropogenic carbon dioxide (CO2) emissions while providing sources of sustainable chemicals and fuels."]
                    \item "2" [label="(2, 0, 0) Direct air capture is energy-intensive and costly."]
                    \item "3" [label="(0, 0, 0) If direct air capture is costly and energy-intensive, then alternative methods are needed for effective carbon capture."]
                    \item "4" [label="(0, 0, 0) Currently, direct air capture is costly and energy-intensive."]
                    \item "14" [label="(0, 0, 0) Deduction-reasoning: Given that direct air capture is costly and energy-intensive, therefore alternative methods are needed for effective carbon capture."]
                    \item "1" -\textgreater "3" [label="deduction-rule"]
                    \item "2" -\textgreater "4" [label="deduction-case"]
                    \item "3" -\textgreater "14" [label="deduction-rule"]
                    \item "4" -\textgreater "14" [label="deduction-case"]
                \end{itemize}

                \item \textbf{Inductive reasoning for the potential of seawater as a carbon sink:}
                \begin{itemize}[nosep, leftmargin=*]
                    \item "5" [label="(3, 0, 0) Dissolved inorganic carbon in seawater has a bicarbonate molarity that is approximately 140 times higher than the 420 ppm of atmospheric CO2."]
                    \item "6" [label="(4, 0, 0) Seawater is a natural carbon sink of net approximately 0.4 giga-ton CO2 per year via the flux exchange between the seawater and atmosphere."]
                    \item "7" [label="(0, 0, 0) Seawater has a high bicarbonate molarity and acts as a natural carbon sink."]
                    \item "8" [label="(0, 0, 0) Induction-reasoning: Given that seawater has a high bicarbonate molarity and acts as a natural carbon sink, therefore it has potential for large-scale CO2 capture."]
                    \item "5" -\textgreater "7" [label="induction-case"]
                    \item "6" -\textgreater "7" [label="induction-case"]
                    \item "7" -\textgreater "8" [label="induction-common"]
                \end{itemize}

                \item \textbf{Abductive reasoning for the use of PEC devices:}
                \begin{itemize}[nosep, leftmargin=*]
                    \item "9" [label="(5, 0, 0) The conversion of dissolved carbon in seawater into carbon-based fuels or chemical products can be entirely powered by sunlight using photoelectrochemical devices."]
                    \item "10" [label="(6, 0, 0) Solar-powered chemical devices floating on the ocean could utilize ocean current, tidal energy, and sunlight to generate dissolved CO2 on demand via bicarbonate acidification."]
                    \item "11" [label="(0, 0, 0) Photoelectrochemical devices can convert dissolved carbon using sunlight, ocean currents, and tidal energy."]
                    \item "12" [label="(0, 0, 0) Abduction-reasoning: If photoelectrochemical devices can convert dissolved carbon using sunlight, ocean currents, and tidal energy, then they are suitable for sustainable CO2 conversion."]
                    \item "9" -\textgreater "11" [label="abduction-phenomenon"]
                    \item "10" -\textgreater "11" [label="abduction-phenomenon"]
                    \item "11" -\textgreater "12" [label="abduction-knowledge"]
                \end{itemize}

                \item \textbf{Final reasoning for proposing the research idea:}
                \begin{itemize}[nosep, leftmargin=*]
                    \item "13" [label="(0, 0, 0) Deduction-reasoning: Given that alternative methods are needed for effective carbon capture, seawater has potential for large-scale CO2 capture, and PEC devices are suitable for sustainable CO2 conversion, therefore developing PEC devices for CO2 conversion in seawater is a viable research direction."]
                    \item "14" -\textgreater "13" [label="deduction-case"]
                    \item "8" -\textgreater "13" [label="deduction-case"]
                    \item "12" -\textgreater "13" [label="deduction-case"]
                \end{itemize}
            \end{enumerate}
        \end{minipage}%
    }
    \caption{A bad RLT Example.}
    \label{fig:dot_graph}
\end{figure*}

\begin{figure*}[t]
    \centering
    \fcolorbox{black}{lightgray!20}{%
        \begin{minipage}{0.9\textwidth}
        \small 

            \textbf{Better Example: DOT Graph Content with Well-Structured Reasoning (Part 1)}

            \begin{enumerate}[nosep, leftmargin=*]
                \item \textbf{Leaf information nodes (no incoming edges):}
                \begin{itemize}[nosep, leftmargin=*]
                    \item "n1" [label="(1,0,0) Single-crystal optical actuators are urgently sought for diverse photonic technologies (molecular machinery, microrobotics, data-storage, quantum-computing)."]
                    \item "n2" [label="(14,0,0) If a [RuSO2] complex achieves 100 
                    \item "n3" [label="(5,0,0) Single-crystal optical actuation has already been demonstrated in many organic, inorganic and organometallic single crystals – hence the phenomenon is chemically general."]
                    \item "n4" [label="(15,0,0) Complete $\eta$1-OSO photoconversion has so far been reported for only two [RuSO2] crystals and neither shows macroscopic actuation – i.e. the desired combination has not yet been realised."]
                    \item "n5" [label="(6,0,0) If a coordination complex contains Ru showing MLCT it will absorb broadly in the visible region and remain thermally robust (favourable for optical actuators)."]
                    \item "n6" [label="(8,0,0) We are synthesising a series of Ru-tetraammine complexes that undergo SO2-linkage photo-isomerisation (the [RuSO2] series)."]
                    \item "n7" [label="(12,0,0) If large photo-induced strain builds up inside a crystal, the macroscopic crystal can bend, peel, crack, fracture or explode."]
                    \item "n8" [label="(19,0,0) Light converts 1 completely at low T and produces substantial micro- and nano-scale strain that disappears on warming."]
                    \item "n9" [label="(17,0,0) The newly discovered crystal 1 actually cracks in a thermally reversible fashion under light (a form of macroscopic optical actuation)."]
                    \item "n10" [label="(10,0,0) If a crystal hosts metastable $\eta$1-OSO or $\eta$2-(OS)O states that thermally revert to the $\eta$1-SO2 dark state, the material can work as a reversible molecular switch."]
                    \item "n11" [label="(22,0,0) The dark $\eta$1-SO2 state of 1 is recovered simply by warming, confirming thermal reversibility."]
                    \item "n12" [label="(23,0,0) Showing three distinct 100 
                    \item "n13" [label="(18,0,0) Crystal 1 gives a pure $\eta$1-OSO structure at 90 K and a pure $\eta$2-(OS)O structure at 100 K, each reached by 100 
                    \item "n14" [label="(24,0,0) In-situ photocrystallography, optical spectroscopy and AFM collectively guarantee that the structural/optical claims about 1 are experimentally secure."]
                \end{itemize}

                \item \textbf{First-level reasoning nodes:}
                \begin{itemize}[nosep, leftmargin=*]
                    \item "A1" [shape=box,label="(0,0,0) Abduction – Photonic technologies need materials whose integer-encoded optical states are fully addressable; therefore we should design single-crystal actuators that can reach 100 
                    \item "I1" [shape=box,label="(0,0,0) Induction – From (i) widespread actuation (n3) and (ii) the absence of examples that unite complete photo-conversion with macroscopic motion (n4), exploring new [RuSO2] derivatives is likely to yield the missing combination."]
                    \item "D1" [shape=box,label="(0,0,0) Deduction – The synthesised Ru-tetraammine complexes (n6) contain Ru with MLCT (n5); hence they should possess broad visible absorption and good thermal stability – desirable traits for optical actuators."]
                \end{itemize}
            \end{enumerate}
        \end{minipage}%
    }
    \caption{Better RLT Example (Part 1): Well-structured DOT graph with leaf information and first-level reasoning.}
    \label{fig:better_dot_graph_part1}
\end{figure*}

\begin{figure*}[t]
    \centering
    \fcolorbox{black}{lightgray!20}{%
        \begin{minipage}{0.9\textwidth}
        \small 

            \textbf{Better Example: DOT Graph Content with Well-Structured Reasoning (Part 2)}

            \begin{enumerate}[nosep, leftmargin=*]
                \item \textbf{Added integrative reasoning nodes:}
                \begin{itemize}[nosep, leftmargin=*]
                    \item "SeriesMerit" [shape=box,label="(0,0,0) Induction – Because exploring [RuSO2] derivatives is promising (I1) and they are predicted to have favourable optical/thermal traits (D1), the new series is a strong candidate for advanced actuators."]
                    \item "HypSeries" [shape=box,label="(0,0,0) Abduction – Given the urgent need for fully addressable single-crystal actuators (A1) and the promise of the new [RuSO2] series (SeriesMerit), synthesising complex 1 is a plausible solution."]
                \end{itemize}

                \item \textbf{Second-level reasoning nodes:}
                \begin{itemize}[nosep, leftmargin=*]
                    \item "CrackConf" [shape=box,label="(0,0,0) Induction – The predicted cracking (CrackPred) and the actually observed cracking (n9) together confirm that 1 genuinely exhibits strain-driven, thermally reversible cracking actuation."]
                    \item "PhysPkg" [shape=box,label="(0,0,0) Induction – Combining the ternary photonic signatures (Ternary) with the reversible-switch behaviour (SwitchRev) yields the conclusion that 1 supplies a multi-state, thermally addressable optical-data package."]
                    \item "FullFeat" [shape=box,label="(0,0,0) Induction – Because 1 unites (i) the multi-state photonic package (PhysPkg) and (ii) confirmed cracking actuation (CrackConf), it simultaneously offers digitally precise optical states and a macroscopic mechanical response."]
                    \item "Validated" [shape=box,label="(0,0,0) Deduction – The experimental triad (n14) and the full feature-set (FullFeat) together prove that the properties of 1 are firmly established."]
                \end{itemize}

                \item \textbf{ROOT: final scientific proposal:}
                \begin{itemize}[nosep, leftmargin=*]
                    \item "ROOT" [shape=ellipse,label="(0,0,0) Abduction – Since a rigorously validated single crystal (Validated) perfectly matches the urgent photonic need via the promising [RuSO2] strategy (HypSeries), we propose trans-[Ru(SO2)(NH3)4(4-bromopyridine)](tosylate)2 (1) as a new single-crystal optical actuator coupling high-purity ternary photonic signatures with thermally reversible cracking motion."]
                \end{itemize}

                \item \textbf{Edges with explicit reasoning types:}
                \begin{itemize}[nosep, leftmargin=*]
                    \item n1 -\textgreater A1 [label="abduction-phenomenon"]
                    \item n2 -\textgreater A1 [label="abduction-knowledge"]
                    \item n3 -\textgreater I1 [label="induction-case"]
                    \item n4 -\textgreater I1 [label="induction-common"]
                    \item n5 -\textgreater D1 [label="deduction-rule"]
                    \item n6 -\textgreater D1 [label="deduction-case"]
                    \item I1 -\textgreater SeriesMerit [label="induction-case"]
                    \item D1 -\textgreater SeriesMerit [label="induction-common"]
                    \item A1 -\textgreater HypSeries [label="abduction-phenomenon"]
                    \item SeriesMerit -\textgreater HypSeries [label="abduction-knowledge"]
                    \item n7 -\textgreater CrackPred [label="deduction-rule"]
                    \item n8 -\textgreater CrackPred [label="deduction-case"]
                    \item CrackPred -\textgreater CrackConf [label="induction-common"]
                    \item n9 -\textgreater CrackConf [label="induction-case"]
                    \item n10 -\textgreater SwitchRev [label="deduction-rule"]
                    \item n11 -\textgreater SwitchRev [label="deduction-case"]
                    \item n12 -\textgreater Ternary [label="deduction-rule"]
                    \item n13 -\textgreater Ternary [label="deduction-case"]
                    \item Ternary -\textgreater PhysPkg [label="induction-case"]
                    \item SwitchRev -\textgreater PhysPkg [label="induction-common"]
                    \item PhysPkg -\textgreater FullFeat [label="induction-common"]
                    \item CrackConf -\textgreater FullFeat [label="induction-case"]
                    \item n14 -\textgreater Validated [label="deduction-rule"]
                    \item FullFeat -\textgreater Validated [label="deduction-case"]
                    \item Validated -\textgreater ROOT [label="abduction-phenomenon"]
                    \item HypSeries -\textgreater ROOT [label="abduction-knowledge"]
                \end{itemize}
            \end{enumerate}
        \end{minipage}%
    }
    \caption{Better RLT Example (Part 2): Well-structured DOT graph with integrative reasoning and final proposal.}
    \label{fig:better_dot_graph_part2}
\end{figure*}

\end{document}